# DEVELOPMENT OF A CARGO SCREENING PROCESS SIMULATOR: A FIRST APPROACH

**Peer-Olaf Siebers [a], Galina Sherman [b], Uwe Aickelin [c]**

[a] Computer Science, Nottingham University, Nottingham, NG8 1BB, UK
[b] CASS Business School, City University London, London, EC1Y 8TZ, UK
[c] Computer Science, Nottingham University, Nottingham, NG8 1BB, UK

[a]pos@cs.nott.ac.uk, [b]galina.sherman.1@cass.city.ac.uk, [c]uxa@cs.nott.ac.uk

**ABSTRACT**
The efficiency of current cargo screening processes at sea and air ports is largely unknown as few benchmarks exists against which they could be measured. Some manufacturers provide benchmarks for individual sensors but we found no benchmarks that take a holistic view of the overall screening procedures and no benchmarks that take operator variability into account. Just adding up resources and manpower used is not an effective way for assessing systems where human decision-making and operator compliance to rules play a vital role. Our aim is to develop a decision support tool (cargo-screening system simulator) that will map the right technology and manpower to the right commodity-threat combination in order to maximise detection rates. In this paper we present our ideas for developing such a system and highlight the research challenges we have identified. Then we introduce our first case study and report on the progress we have made so far.

## 1. INTRODUCTION

The primary goal of cargo screening at sea ports and air ports is to detect human stowaways, conventional, nuclear, chemical and radiological weapons and other potential threats. This is an extremely difficult task due to the sheer volume of cargo being moved through ports between countries. For example in sea freight, 200 million containers are moved through 220 ports around the globe every year; this is 90% of all non bulk sea cargo (Dorndorf, Herbers, Panascia, and Zimmermann 2007).

Little is known about the efficiency of current cargo screening processes as few benchmarks exist against which they could be measured (e.g. %detected vs. %missed). Some manufacturer benchmarks are available for individual sensors, but these have been measured under laboratory conditions. It is rare to find unbiased benchmarks that come from independent field tests under real world conditions. Furthermore, we have not found any benchmarks that take a holistic view of the entire screening process assessing a combination of sensors and also taking operator skills, judgment and variability into account.

In our research we attempt to identify and test innovative methods in order to advance the use of simulation for supporting decision making at the strategic and the operational level of the cargo screening process. Wilson (2005) confirms the usefulness of simulation for the analysis and prediction of operational effectiveness, efficiency, and detection rates of existing or proposed security systems.

Our research aim is to develop a methodology for building such Decision Support Systems (DSS) that will map the right technology and manpower to the right commodity-threat combination in order to maximise detection rates. The concept for such a DSS (a cargo screening process simulator) is shown in Figure 1. For developing the methodology we are using a case study approach. In our work we focus solely on DSSs development; we do not work on new sensor development. However, with our DSSs we might be able to give some recommendations of what characteristics new to be developed sensors might require to reach certain system performances.

The core of the proposed cargo screening process simulator will consist of three elements: a Detection Rate Matrix (DRM), a simulation model and a resource optimiser. The DRM will provide sensor detection rates as an input for the sensors represented in the simulation model, based on sensor types, commodities, threats, and other indicators. The simulation model will allow carrying out what-if analyses for the system under examination. The results of the simulation will be fed into the resource optimiser to create a new set of input parameter values for the simulation. The previous two steps are repeated until an acceptable solution has been found. The output of the simulator will consist of required technology and manpower and an estimation of the system detection rate that can be achieved by implementing the proposed system set-up. A sensor data database will provide some information for the core elements (in particular for the DRM). The content of the database will be a mixture of data provided by vendors but will also consider operators experience with the equipment. Other input data required for the cargo screening process simulator include an annual job list, guideline on how to carry out jobs, and observations if jobs are carried out in accordance with these guidelines, and a list of existing sensors and staff.

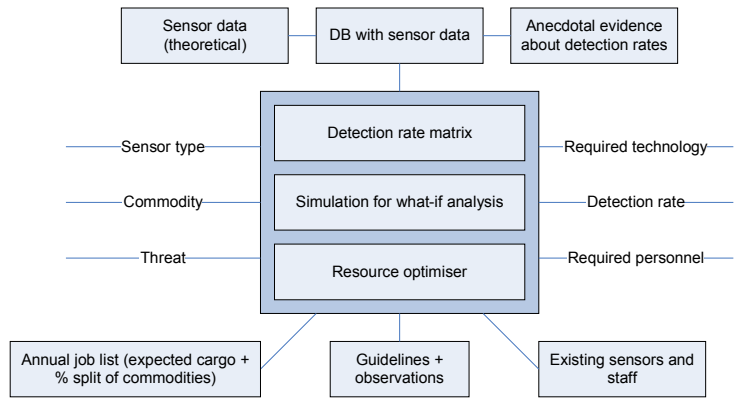

Figure 1: Conceptual Model of Our Cargo Screening Process Simulator

Section 2 contains a brief review of existing work in the field. In Section 3 we discuss the development of a DRM. In Section 4 we state our research questions regarding model design and matrix development. Section 5 introduces our case study, the ferry port in Calais. We present a description of the real system and its operations, a conceptual model of it, an implementation of the conceptual model in form of a discrete event simulation model, and finally we show the results of an initial test run with our simulation model. Section 6 concludes the paper by discussing the results of our current efforts and proposes further work.

## 2. BACKGROUND

Simulation modelling is commonly used to support design and analysis of complex systems. With regards to modelling ports Tahar and Hussain (2000) confirm that simulation modelling is a tool widely used for the management, planning and optimisation of port systems. According to Turner and Williams (2005) the same is true for the management, planning and optimisation of complex supply chain systems.

In the context of the cargo screening process, some examples (e.g. Leone and Liu 2005, Wilson 2005) have been found that use simulation modelling to evaluate key design parameters for checked baggage security screening systems in airports, in order to balance equipment cost, passenger and baggage demand, screening capacity, and security effectiveness in an attempt to meet the requirements imposed by the checked baggage screening explosive detection deadline established by the US Aviation and Transportation Security Act.

Another related subject is the enhancement of the security throughout the supply chain, i.e. achieving supply chain integrity (Closs and McGarrell 2004). Here, simulation modelling is often used to analyse the system. For example, Sekine, Campos-Náñnez, Harrald, and Abeledo (2006) use simulation and the response surface method for a trade-off analysis of port security in order to construct a set of Pareto optimal solutions. The development of a dynamic security airport simulation is described by Weiss (2008). In contrast to the other papers mentioned so far, this simulation focuses on the human aspects in the system and employs the agent paradigm to represent the behaviour of attackers and defenders. Both, attacker and defender agents are equipped with the capability to make their individual decisions after assessing the current situation and to adapt their general behaviour through learning from previous experiences. This allows accounting for rapid security adaptation to shifting threads, as they might be experienced in the real world.

## 3. CONCEPTS OF THE DETECTION RATE MATRIX

The mapping process (right technology and manpower to the right commodity-threat combination) will be implemented using a multi-dimensional DRM. The DRM contains the information required to estimate the type and amount of sensors and manpower we need in order to maximise our detection rate if we have an estimate of the number and type of cargo containers we want to screen and what they will contain. The values to fill the DRM can either come from vendors, the literature, from trials, or anecdotal evidence of the border agency staff. From all the information received we have to create a single value that represents the detection rate for a certain commodity-threat combination.

An example for a partially filled DRM derived from laboratory experiments can be found in Klock (2005). Klock states that developing a DRM from real world data would be desirable but poses a big challenge as for various reasons it is a problem to collect all relevant data for the all commodity-threat combinations in the real system. In our case the problems are as follows. In most cases the security screening procedures cannot be compromised for research purposes, i.e. there are legal boundaries regarding the sampling frame. Furthermore, it would be difficult to capture the variability of operational procedures that exist in the real system. However, as much as the technology itself, the way in which the technology is used contributes to the success rate of detecting threats. Our current plan is to fill some of the gaps in our DRM by simulating specific scenarios, rather than trying to collect all data from the real system.

We will start the development of our DRM by creating a two dimensional matrix and then gradually increase its complexity (i.e. the number of dimensions). The values for our first DRM will be derived by collecting anecdotal evidence from system insiders and where anecdotal evidence is not available by simulating specific scenarios of interest (1). The next step will be to generalise the initial DRM and to consider that the applicability and performance of sensors is related to the commodity screened and the category of threats investigated (2). For example, if one wants to detect stowaways in a lorry using CO2 probes which measure the level of carbon dioxide and the load consist of wood or wooden furniture which naturally exhumes carbon dioxide then the detector readings will be wrong. For this commodity the sensor is not useful and would produce many false positives (type 1 error), which means that in return many false negatives (type 2 error) will stay undetected as time is wasted with manually inspecting the wrong lorries. The next dimension we will add is a definition of the cargo containment which consists of a description of the type of containment, its wall thickness and its wall density (3). The containment type is important as some of the sensors might need to have access to the interior of the containment while others might be applicable to be used from the outside. Wall thickness and density are important as many sensors have limitations regarding the penetration of the containments, depending on the containment properties.

rate of detection = f (commodity & thread combination, specific scenario)   (1)

rate of detection = f (commodity, threat, sensor)   (2)

rate of detection = f (cargo containment, commodity, threat, sensor)   (3)

There are many more dimensions one could add (e.g. cargo origin, cargo destination, shipping company, or environmental conditions of test facility location) and part of the research will have to deal with the question of which are the most relevant indicators of sensor efficiency?

## 4. RESEARCH QUESTION

One of the key questions we are keen on answering during our research is how and where it makes sense to use simulation in a project like ours. Besides the standard application areas for simulation modelling in operations research (e.g. system analysis, optimisation, as a communication tool) we want to find and test some new application areas (e.g. validating the DRM parts where we have data and helping to estimate the values where we have gaps in our DRM). Furthermore, we will examine if our simulation models can be used to support the decision making process in other fields, e.g. supply chain management or risk analysis.

Before we can build our cargo screening process simulator we will have to investigate several questions which can be broadly grouped in two categories, related to model design or matrix development. Research questions regarding model design: [a] How much detail do we have to model to get some meaningful output? [b] How should we model people in our system (e.g. officers or stowaways) - as simple resources or as autonomous entities? [c] How can we get a good estimate on how many stowaways, weapons or drugs are passing the borders? [d] What effect does the fact that we are dealing with rare events have on input sampling and output analysis? Research questions regarding matrix development: [e] Which are the most relevant indicators of sensor efficiency? [f] What is the best way to develop and validate a detection rate matrix in absence of real data or when real data is incomplete, i.e. missing data for certain technology / commodity / threat combinations? [g] Can we develop a framework to support the development of a DRM for different environments and for different threats?

## 5. CASE STUDY: CALAIS FERRY PORT

In order to achieve our research aim of developing a methodology for building cargo screening process simulator we have chosen to use a case study approach. This allows us to gain the knowledge, insight, and experience we need for developing our methodology. For each case study we will first develop simulation models that allow us to analyse the system under study and then create a DRM for this system.

For our first case study we have selected the ferry port in Calais (France) that links Calais with Dover (UK). This site is ideal for beginning as the security measures in place focus on detecting only one threat, illegal immigrants, or clandestines, as they are called by the UK Border Force. Clandestines are people found on a lorry with the aim to get into Britain without a passport or any other papers (Sky1 2009). These can be individuals or groups. Clandestines come in hope of a better future in Britain, drawn by the English language, the lack of national identity cards and the possibility of illegal work. When clandestines do not succeed little or no publicity is generated, thereby perpetuating the false idea that clandestines are always successful. On the other hand, for every successful clandestine arriving in Britain the word goes out that the process is successful, which generates even more attempts of illegal immigration (Brown 1995).

### 5.1. The Real System

Between April 2007 and April 2008 more than 900,000 lorries passed the check points in Calais. Of these, approx. 0.3% contained additional human freight (UK Border Agency 2009). How many clandestines were missed during these checks is unknown. Although companies supplying the sensor technology promise a detection rate close to 100%, independent test have shown that this is not the case when using the equipment in real world scenarios (Klock 2005). In addition, in the real system the detection rates also depend on factors like the time of day (at busy times the operators have less time to apply the sensors and wait for the readings and therefore readings are more likely to produce more type I and type II errors), operators' skills (of interpreting the outputs from the sensors), and operators' fatigue.

In Calais the cargo screening process is separated into two major zones, the first under the control of the Calais Chamber of Commerce (CCI), the second under the control of the UK Border Agency (see Figure 2). Different types of sensors are used at the various screening facilities and some of them are also in use as mobile devices. The technology / operations used for screening includes Passive MilliMetre Wave scanners (PMMW), Heart Beat Detectors (HBD), $CO_2$ measurement probes (CO2), canine sniffers and visual inspection. The process on the French site starts with a passport check by the French authorities. Then all lorries are screened for clandestines and suspicious lorries are routed to deep search facilities where they are further inspected by using an alternative method and if suspicion is substantiated then lorries are opened for visual inspection. In some cases (e.g. if it does not interrupt the process flow much, e.g. at non-busy times) lorries are opened directly for a quick visual check after or instead of being screened. If clandestines are found on board a lorry they are removed by the French police, registered, and released into freedom. The process on the UK site is very similar; the major difference is that lorries are searched rather then screened and that only a fraction of the lorries going through the system is actually searched (at average 33%). The number of vehicles searched is on the basis of profiling and intelligence. Once the lorries have passed all check points they park at the Berth where mobile squads are operating to check the lorries a last time before they get on their way to Dover.

**5.2. Modelling the Real System: A First Approach**
This initial modelling exercise acts as a data requirement analysis for our case study. It will help us to make informed decisions about the information and data we need to collect during our main data collection for this case study. Furthermore, it will help us to uncover areas where we might encounter problems during our main modelling and implementation process at an early stage, so that we can respond to it in sufficient time. Finally, we want to use our initial models to communicate theories, ideas, potential investigation techniques, outputs and solutions to stake holders and other interested parties.

Before we started our modelling exercise we visited the case study site to observe the operations, for discussions with stake holders, and for collecting system performance data. From the information gathered we developed a conceptual model that reflects the current operations of the cargo screening process at the ferry port in Calais.

**5.2.1. Modelling Challenges**
The case study system presents several modelling challenges, some of which have already been mentioned in Section 4. Below is a list of the modelling challenges we are currently facing. The first challenge is related to the fact that we are dealing here with a complex system where factors that are difficult to quantify are assumed to have a big impact on system behaviour and ultimately system performance. An example for such a factor is the human decision making process. Therefore, the application of abstraction and simplification for the purpose of model design is a very delicate issue.

The second challenge is related to the lack of input data. On the one hand we are dealing with rare events (e.g. detecting a clandestine) which impacts on the way we have to do our input sampling and output analysis (Heidelberger 1995) and some data cannot be obtained from the real system (e.g. number of clandestines that manage to cross the borders) so we have to make a lot of guesses. Some mathematical models exist for estimate such values as for example success rates for clandestine border crossing (Wein, Liu, and Motskin 2009; Epenshade 1995); their usefulness however is debatable as still many assumptions have to be made to derive these estimates. Even if we had the resources to collect the data there are some legal issues regarding the sampling frame which prohibits us to collect some of the required data as we are not allowed to sample an entire population.

The third challenge relates to the objects we have to model, some of which are fixed and some of which can be moving around freely. Three different situations can be identified: [1] Sensors are fixed and targets (lorries and clandestines) are fixed, for example in a screening shed lorries are parking while sensors are applied. [2] Sensors are moving and targets (clandestines) are moving, for example in the allocation lanes officers are patrolling and clandestines are running around in order to get into the lorries. [3] Sensors are moving and targets (lorries and clandestines) are fixed, for example in the Berth squads are checking the parking lorries either with mobile sensors or by opening suspicious lorries directly. While the first situation is relatively easy to model using traditional Discrete Event Modelling (DEM) the latter two require some further reflection before they can be modelled successfully. In those cases sensors and/or targets need to possess some form of autonomy and probably proactiveness which are concepts not directly supported by traditional DEM. Agent-Based Modelling (ABM) presents an alternative modelling paradigm that supports the consideration of autonomy and proactiveness of entities.

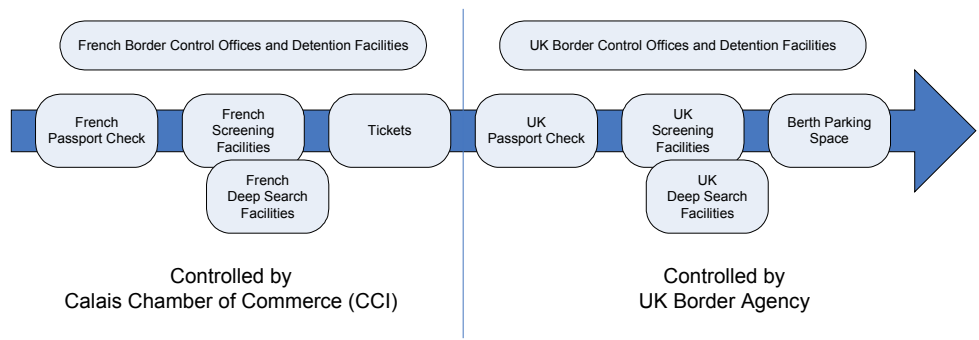

Figure 2: The cargo screening process at Calais

The fourth challenge relates to the injection of clandestines into the model. Anecdotal evidence suggests that clandestines enter the system at numerous places. Clandestines either get into the lorries before these enter the compound or they climb over the fences that surround the compound and get into the lorries while they pass through the compound or wait for the ferry. While the first is easy to model the latter causes some problems as no data is available exactly where and when the clandestines enter the compound.

Finally, the fifth challenge relates to the human decision making. The operation of this system is human centric and relies very much on the experience of the officers and the compliance to rules. Human decision making involves the routing (i.e. choosing the lorries to be screened), choosing the sensor to be used, interpreting the sensor outputs (i.e. choosing the lorries to be opened), and compliance to rules (sticking to recommended sensor application periods). All these points depend very much on the state of the system. At peak times decisions will be different compared to quiet times, e.g. sensor application periods will be shorter to avoid congestion in front of the service sheds and therefore the number of true and false negatives will be much higher and therefore the detection rates vary throughout the day.

In the end the big question is if modelling all these details is really necessary for getting useful results. In order to answer this question we will have to implement them at least partially and conduct a sensitivity analysis. For this purpose we will build some smaller simulation models that only represent a small section of the overall real system. Once we have the results we can give some recommendations regarding the level of detail that is required for getting a useful representation of the real system.

### 5.2.2. The Conceptual Model

In order to capture the cargo screening process taking place at the Calais ferry port we have developed a process centric conceptual model (Figure 3). It reflects the process flow as it appears in the real system. Dark blue fields represent system entry and exit points. Brown fields represent jump starting points (where the text is followed by @) and targets (where the text is lead by @). These jumps do not consume any time. Light blue fields represent the locations where time is consumed. The %s represent flow probabilities while the numbers below the light blue fields represent detection rates. For confidentiality reasons the true values have been replaced by place holders. The splits in the model have been defined in a somewhat arbitrary way but often a single row represents all activities that happen at one specific location.

### 5.3. Implementation

Based on our conceptual model presented in Section 5.2.2 we have developed a first version of a Discrete Event Simulation (DES) model which is implemented in AnyLogic Version 6.4, a java-based multi-paradigm simulation software. The purpose of this exercise is to identify where we have gaps in knowledge about the system and to identify missing data that could be obtained during our main data collection.

### 5.3.1. Simulation Software

The object-oriented model design paradigm supported by AnyLogic provides for modular, hierarchical, and incremental construction of large models (XJ Technologies 2009). Each model contains a set of active objects which often represent objects found in the real world. At the lowest level these active objects can contain parameters, variables, functions, events, state charts and other active objects. For DES modelling there is also a library containing higher-level objects that support the creation of discrete event patterns frequently used in process-centric modelling (e.g. entity generation, buffering, resource usage, entity routing, entity destruction).

One of the benefits of AnyLogic is that you can build mixed models, i.e. you can mix process-centric DEM and individual based ABM in one hybrid simulation model. Technically the main difference is that an agent compared to an active object has some additional features with respect to dynamic creation and destruction, synchronisation, space-, mobility-, and spatial animation, agent connections and agent communication. We will use these features when we model for example sensor and target movement in the allocation lanes.

For our current simulation model we use the elements from the library but we have also developed our own element in form of embedded active objects that contain a collection of parameters, variables and library elements. These are reusable components that can currently represent any of the service sheds as well as passport and ticket booth on the Calais compound.

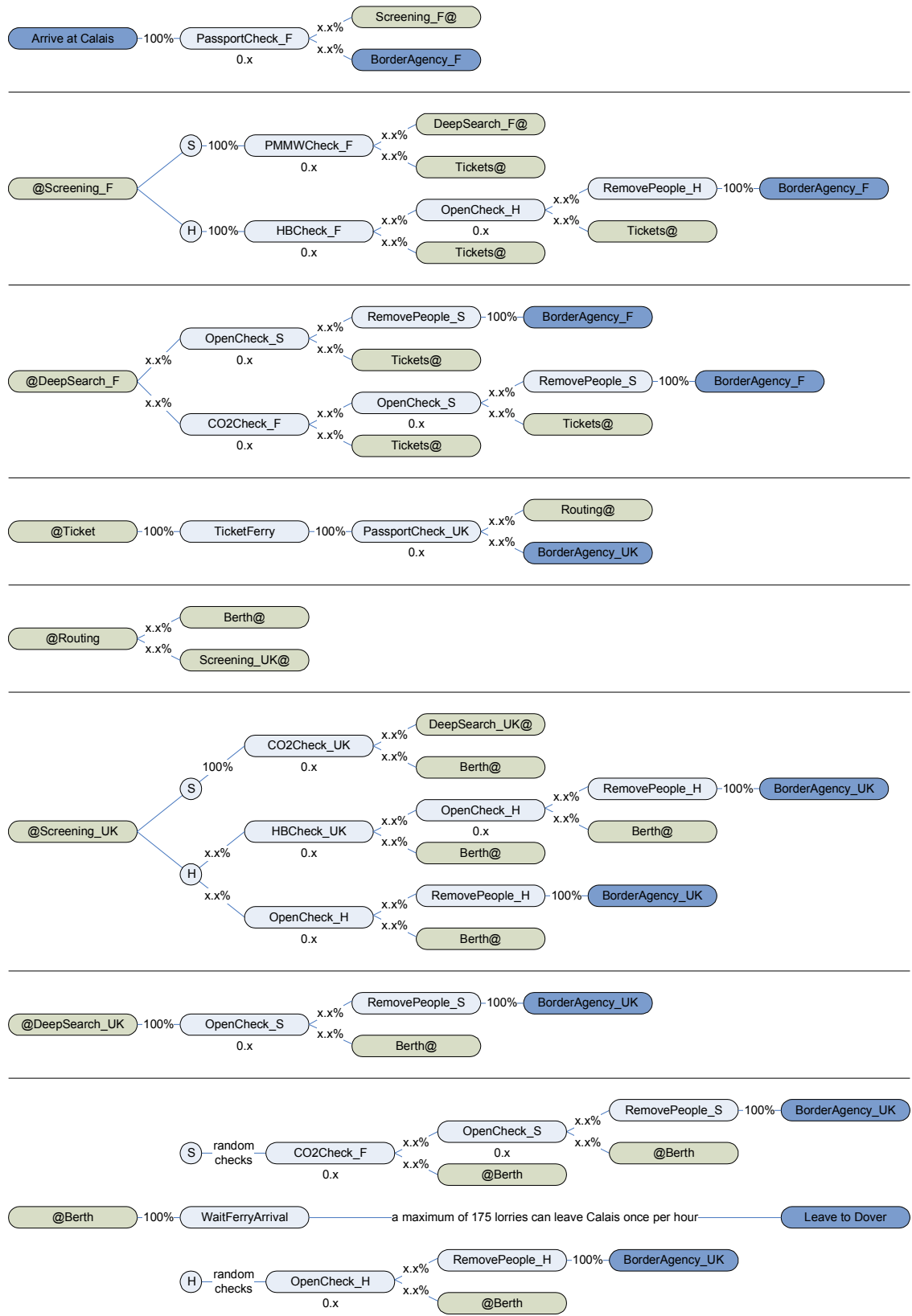

Figure 3: Conceptual model of the cargo screening process at Calais (S=soft sided, H=hard sided, F=French site; UK=UK site; x=replacement for real value)

We tried to make them as generic as possible so that we can also use them for modelling other locations or types of systems. A screenshot of the elements of such an embedded active object is presented in Figure 4. It shows a service station with a linked resource pool (symbolised by the clock and the linked box) an entrance buffer and two single space exit buffers. The hold element between entrance queue and service station is released to let one entity pass at a time as soon as the previously serviced entity has left the exit buffer, which only happens if there is some space available in one of the upstream queues. The variables on the left are used for data collection while the parameters on the right allow each instance of this active object class to be defined by an individual set of parameters.

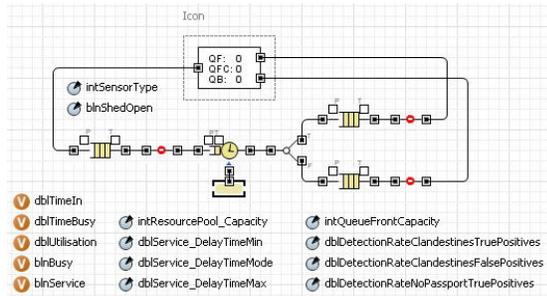

Figure 4: Embedded active object serviceShed

### 5.3.2. Simulation Model

Entities of type Lorry (soft and hard sided) are injected into the simulation at a certain rate by the source element (arrival). Some lorries will arrive with an additional load (clandestines) on board. Clandestines are currently modelled as resources (a boolean variable defines if there are clandestines on board a lorry or not). The main elements in the simulation model are the serviceShed elements which have been described in Section 5.3.1. These are used for modelling the situations when we have fixed sensors and fixed targets. The serviceShed elements are linked via some routing elements. The routing elements use a custom-made function which routes the entity to the next level upstream element with the shortest queue. This represents the routing activities normally conducted by an officer.

Figure 5 displays a section of the simulation model within the AnyLogic IDE. The project view window on the left shows the project tree of the current project. The graphical editor in the middle shows the content of the Main object. The pallet window on the right displays the different pallets available in AnyLogic, amongst them the Enterprise Library pallet. The properties window at the bottom is used to define the properties of the element, which can contain Java commands and method calls.

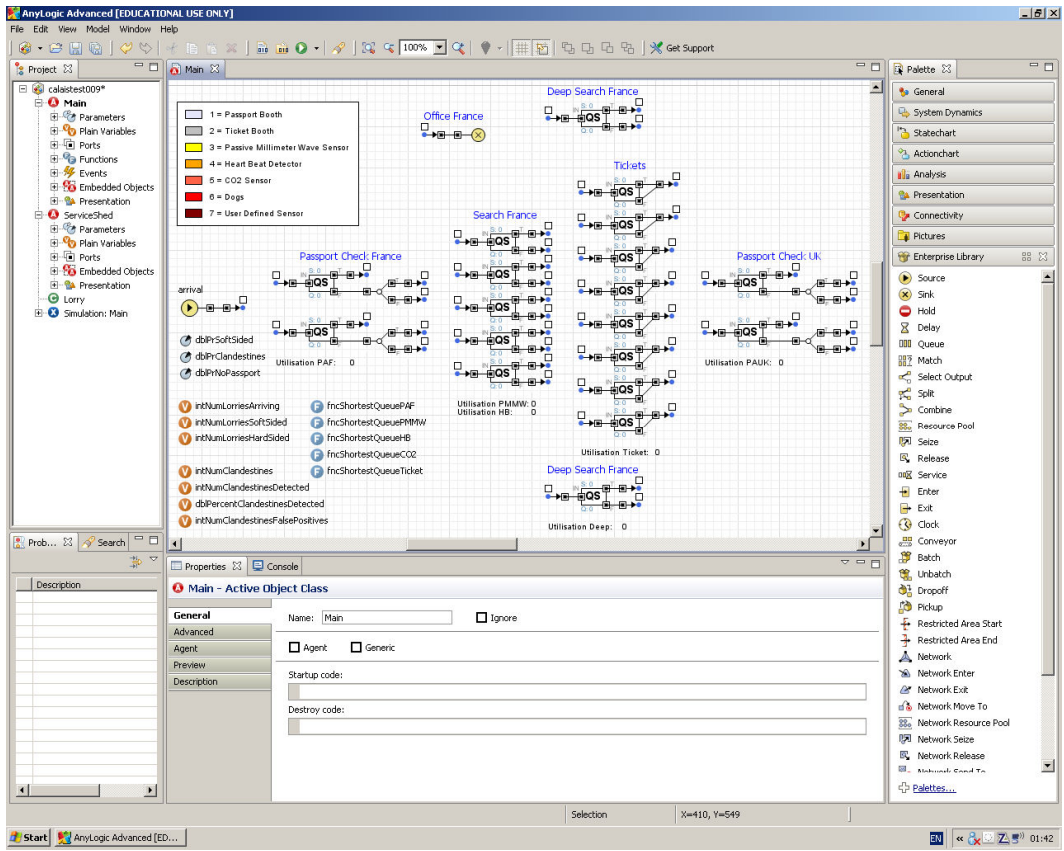

Figure 5: A section of the simulation model within the AnyLogic IDE

For modelling the Berth activities we could not use our serviceShed elements as we have a situation with moving sensors and fixed targets. Instead we developed the solution presented in Figure 6. The moving squads are modelled by events that pick one lorry at random (mobile $CO_2$ checks will be conducted on soft sided lorries while hard sided lorries will be opened) and check it. The time it takes to check a lorry is represented by the inter arrival time between two events. This means that the squads are currently modelled as being 100% utilised as long as there are lorries to check. There are two modus operandi, either lorries can be checked only ones (lorries that have been checked already are registered on an ignore list) or lorries can be checked multiple times (which represents the situation where clandestines enter the lorries while these are parking at the Berth and therefore the squad would check suspicious lorries again).

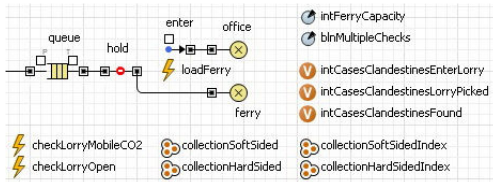

Figure 6: Modelling the Berth activities

To set up the model parameters we had to use some best guesses and common sense. However, as we have some average input and output data we have tried to use settings for the unknown parameters to match the output data of the real system when using the input data of the real system. A problem is that the available data are average data for multiple stations or monthly averages. Where ever possible we have used multiple data sources for estimating values for the data required (e.g. for calculating process flow probabilities).

### 5.3.3. Current Omissions

This simulation model presented here represents the first draft of our aspired DES model. The main simplifications and abstractions in the current model are listed below:

- Queues: We use large queue capacities in front of the service stations; therefore congestion does not occur. However, it is an important phenomenon that occurs in the real system and influences service times and therefore indirectly the detection rates.
- Average values: We use the same average entity arrival rates for the entire simulation runtime but the collected data indicates a significant difference in arrival rates as well as inspection and detection rates depending on time of the day and day of the week (the higher the arrival rates the lower detection rates, as officers have less time for conducting an individual screening). However, first we need to sort out the congestion problem mentioned above; otherwise the impact of high arrival rates is not adequately considered in the results.
- Currently we don't model multiple clandestine entry points, canine sniffers nor the search for clandestines in the allocation lanes.

### 5.4. Testing the Simulation Model

So far we have only conducted some very basic preliminary tests with our simulation model. A verification and validation exercise is still to be carried out. However, here we briefly report on one of the tests we have conducted. We have set up the simulation model using our standard set of parameter values, except for the sensor detection rates, which we have set to the same value for all sensors. During the experiment we have systematically changed this collective value, starting from 0% to 100% in steps of 10%. Our simulated runtime was equivalent to a one year period and we conducted 20 replications for each set of values. As for the results we expect to see a non-linear relationship between sensor detection rates and the average proportions of clandestines detected. This is due to the fact that many lorries will go through several screening procedures and therefore combinatorial effects appears for this relationship, where higher individual sensor detection rates will have a proportionally lower benefit regarding the system detection rate. Figure 7 confirms our expectation.

This first test has already shown the impact of modelling rare events. We observed that the clandestine detection rates vary significantly throughout most of the simulation runtime and seems only to stabilise towards the end. Furthermore, we noticed some significant differences between runs. Therefore, in future we have to assess very carefully the required warm-up period, run length and number of replications.

### 6. CONCLUSIONS

In this paper we have presented our first steps towards the development of a cargo-screening process simulator and we have introduced a case study that we want to use for gaining some experience with developing such a simulator. Our current task is to conduct a data requirement analysis. For this we have created a first draft of our aspired DES model to be used for the cargo-screening process simulator. This modelling exercise has allowed us to make a well informed decision about which kind of information and data we require for representing the real system to allow some useful systems analysis.

We found that a big challenge when modelling the case study system is to capture the variability inherent in the system. By omitting details like differences in arrival rates throughout the day and week, congestion in front of service sheds and associated with this service time variation and detection rate variations we do not get a good representation of the real system, in particular when we are not only interested in the average system performance but also want to gain an insight into its operations. We are currently working on the mechanisms to implement varying arrival rate and the consequences of these, i.e. congestion and varying service times. Once we have conducted our main data collection we will add some real data to it. Once this is done we will work on verification and validation of the simulation model.

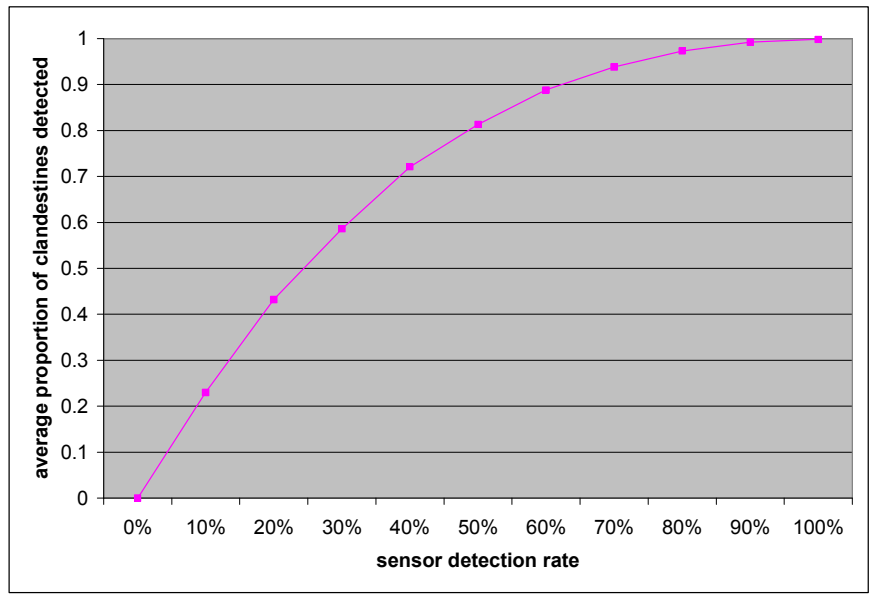

Figure 7: Results from the experiment: sensor detection rates vs. proportion of clandestines detected

## ACKNOWLEDGMENTS

This project is supported by the EPSRC, grant number EP/G004234/1 and the UK Border Agency.

**AUTHORS BIOGRAPHY**

PEER-OLAF SIEBERS is a Research Fellow at The University of Nottingham, School of Computer Science. His main research interest is the application of computer simulation to study human-centric complex adaptive systems. This is a highly interdisciplinary research field, involving disciplines like social science, psychology, management science, operations research, economics and engineering. Furthermore, he is interested in nature inspired computing and agent-based robotics. For more information see http://www.cs.nott.ac.uk/~pos/

GALINA SHERMAN is a PhD student at City University, Cass Business School. Her current research is related to supply chain management, risk analysis and rare event modelling.

UWE AICKELIN is a Professor of Computer Science and an Advanced EPSRC Research Fellow at The University of Nottingham, School of Computer Science. His main research interests are mathematical modelling, agent-based simulation, heuristic optimisation and artificial immune systems. For more information see http://www.cs.nott.ac.uk/~uxa/